\documentclass[letterpaper, 10 pt, conference]{ieeeconf}  

\IEEEoverridecommandlockouts                              

\usepackage{math}   
\usepackage[capitalize]{cleveref}
\crefname{section}{Sec.}{Secs.}
\Crefname{section}{Section}{Sections}
\Crefname{table}{Table}{Tables}
\crefname{table}{Tab.}{Tabs.}

\title{GeoLCR: Attention-based Geometric Loop Closure and Registration}

\author{Jing Liang$^{1}$, Sanghyun Son$^{1}$, Ming C. Lin$^{1}$ and Dinesh Manocha$^{1}$ \\
\url{https://youtu.be/Nr0jFBDD_Pk}
\thanks{$^{1}$ Authors are from Computer Science Department, University of Maryland, College Park}%
\thanks{Website:\url{https://gamma.umd.edu/researchdirections/autonomousdriving/geolcr/}}
\thanks{Code: \url{https://github.com/jingGM/GeoLCR.git}}
}

\begin{document}

\maketitle
\thispagestyle{empty}
\pagestyle{empty}

\begin{abstract}
We present a novel algorithm specially designed for loop detection and registration that utilizes Lidar-based perception. Our approach to loop detection involves voxelizing point clouds, followed by an overlap calculation to confirm whether a vehicle has completed a loop. We further enhance the current pose's accuracy via an innovative point-level registration model. The efficacy of our algorithm has been assessed across a range of well-known datasets, including KITTI~\cite{kitti}, KITTI-360~\cite{kitti360}, Nuscenes~\cite{nuscenes}, Complex Urban~\cite{urban}, NCLT~\cite{nclt}, and MulRan~\cite{mulran}. In comparative terms, our method exhibits up to a twofold increase in the precision of both translation and rotation estimations. Particularly noteworthy is our method's performance on challenging sequences where it outperforms others, and achieves mostly $100\%$ success rate in loop detection.
\end{abstract}

\section{Introduction}

Simultaneous Localization and Mapping (SLAM) algorithms are integral components in robotics and autonomous navigation applications~\cite{fastlio, lego, overlapnet, lcdnet}. These algorithms strive to simultaneously estimate the position of a vehicle navigating within an unknown environment and construct a map of the environment. SLAM computation generally encompasses two primary steps~\cite{tsintotas2022revisiting, lowry2015visual}: The first is local odometry~\cite{gicp, lego, fastlio}, which aims to construct a consistent map within a short range. However, this approach tends to accumulate drift over time. To counteract this, the second stage, loop closure~\cite{elastic, lcdnet, overlapnet}, comes into play when the robot encounters a closed loop, enabling the rectification of the constructed map and addressing drift accumulation issues. Our focus in this paper is only on loop closure, specifically on loop detection and pose computation.

A wide variety of sensors are used in different vehicles, such as mono-cameras, RGBD cameras, Lidars, and IMUs. Different perception algorithms are also developed that are tailored for each sensor. Monocular cameras, for instance, are commonly employed in SLAM tasks~\cite{orbslam}, and the efficacy of visual SLAM can be further optimized with an IMU sensor. The advent of RGBD cameras has enabled these algorithms to compute precise depth data, thereby enhancing the map-building accuracy. Among the sensors, Lidars are often considered the most accurate for long-range depth information. Several approaches~\cite{lego, liosam, lcdnet, overlapnet} exploit Lidars for mapping and localization as they typically provide higher quality depth information compared to RGBD sensors. In this paper, we confine our attention to Lidar as the perceptual sensor for the loop closure tasks.

\begin{figure}
    \centering
    \includegraphics[width=\linewidth]{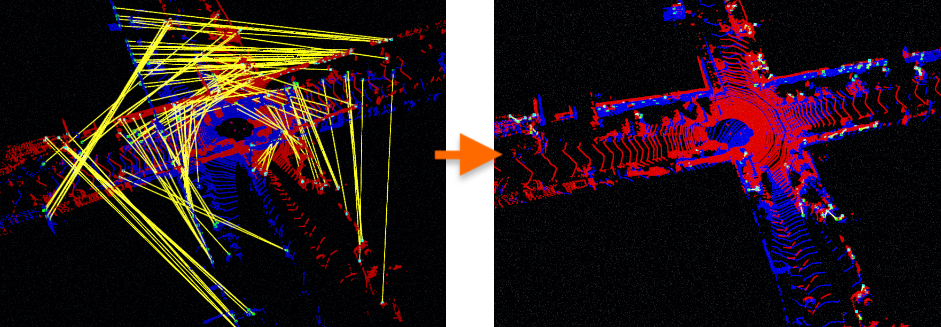}
    \caption{The overlap estimation and registration when the car arrives at a previous location in a cross. The yellow lines on the left represent the matched voxels and the right side is the map after registration.}
    \label{fig:cross-meet} 
\end{figure}

Loop closure in SLAM computation consists of two subtasks, as depicted in Figure \ref{fig:cross-meet}. The first task involves detecting the closed-loop scenario where the robot needs to discern if the current environment resembles any previously encountered scenarios. The second task involves recalculating the current position based on detected similar frames for low-level map rectification. Traditional approaches~\cite{lego, liosam} often use ICP-based methods, yet these optimization-based algorithms often suffer from local minimum issues. Some learning-based approaches~\cite{overlapnet, arshad2021role} target autonomous cars, focusing on yaw angle registration and long-range Lidar points. However, these approaches do not offer distance-related information and cannot be directly implemented by small robots that require 3D rotation and translation with short-range Lidar sensors.

\noindent \textbf{Main Contributions: }  
We present GeoLCR, a learning-based approach for geometric loop closure and point-level registration. The framework involves a unique amalgamation of overlap estimation and registration, incorporating innovative techniques for voxel matching and pose calculations to bolster accuracy. The unique aspects of our work encompass:

\begin{enumerate}
\item We propose an overlap detection metric for 3D loop closure detection. Based on the Superpoint Matching Module~\cite{geometric}, we develop an overlap estimator to estimate the overlap values. We demonstrate that the overlap estimation is consistent with the distances between two point clouds and is easy to use for loop closure detection. 

\item We propose a novel transformer-based registration model that is specific to our overlap estimator. We exploit the fact that the distance between coarse features corresponds to the overlap between a pair of voxels and take these features into account to improve the accuracy of point matching. We also show that our registration transformer generates better results (Section \ref{sec:experiments} and Appendix~\cite{appendix}).

\item We validate the efficacy of our approach using different datasets, including those based on cars and small robots, specifically KITTI~\cite{kitti}, KITTI-360~\cite{kitti360}, Nuscenes~\cite{nuscenes}, Complex Urban~\cite{urban}, NCLT~\cite{nclt} and MulRan~\cite{mulran}. We show that our approach is up to two times more accurate in registration and  also more accurate in overlap detection than the other state-of-the-art methods~\cite{appendix}.
\end{enumerate}

\section{Background}
Different techniques have been proposed for loop closure based on the underlying sensors.  In terms of camera-based and Lidar-based approaches, there are many traditional approaches as well as learning-based approaches~\cite{zhang2021visual, overlapnet, lcdnet}. In this section, we mainly limit ourselves to techniques proposed for Lidar sensors.

\subsection{Loop Closure Detection}
The traditional loop closure algorithms using Lidar data are mostly based on optimization approaches~\cite{lego, fastlio}, where the Levenberg-Marquardt algorithm is used to find the minimum distance between the point clouds directly ~\cite{fastlio, dellenbach2022ct}. 
The Lio-sam~\cite{liosam} algorithm uses an extra GPS sensor and introduces the GPS graph factor to help with the detection of loop closure. LLoam~\cite{ji2019lloam} segments the Lidar data and builds a factor graph to perform graph optimization for loop closure detection. LDSO~\cite{gao2018ldso} uses the image features to build a similar bag-of-words (BoW) approach and uses a sliding window to perform optimization. However, those approaches suffer from local minimum problems, where the current pose estimation may not be accurate. 

With the development of learning-based approaches, researchers are able to extract the semantic information from point clouds and compare the semantic features from different frames for loop closure. The learning-based methods implicitly trained the models to detect the loops comparably accurately or even more accurately than classical methods~\cite{lin2021topology, li2021sa,zhou2022loop}. Lin et al.~\cite{lin2021topology} use the semantic information to build a topology graph and compare the objects' relative positions within two frames; SA-LOAM~\cite{li2021sa} extracts the semantic information of each sub-graph (i.e. the local map) and then compares the similarity of the sub-graphs for loop closure. \cite{zhou2022loop} uses a similar approach to generate local map descriptors and compare the likelihood of the descriptors. Although these semantic approaches can compute the similarities of different local frames, they cannot predict the overlap or matching of different local frames. Recently, several approaches have been proposed that can directly predict the overlap between the local frames~\cite{zhou2021ndt, ma2022overlaptransformer, overlapnet}. However, their predictions are based on the 2D birds-eye-view graph, and they directly use the overall point cloud for registration, which can be computationally expensive and requires a lot of storage. In this paper, we present a new method based on a Superpoint Matching Module~\cite{geometric} to predict the overlap and use the points in matched voxels to perform point registration and predict the current position.

\subsection{Point Registration}
After detecting the loop closure, a new estimate of the current position needs to be predicted with the matched frame. There are many known approaches that can be used for point registration. Traditionally, researchers have used ICP/GICP~\cite{gicp} for registration with the help of RANdom SAmple Consensus (RANSAC). For learning-based approaches, the Sinkhorn algorithm and differentiable Singular Value Decomposition (SVD) approaches have been combined for the registration task~\cite{geometric, lcdnet}. These approaches take into account all the points for calculation, which can be computationally costly and subject to local minima issues.  We present a novel approach using a point transformer that takes the features of matched voxels for registration and estimates the transformation matrix between the points in matched pairs to reduce the computation.


\begin{figure*}
    \centering
    \includegraphics[width=0.85\linewidth]{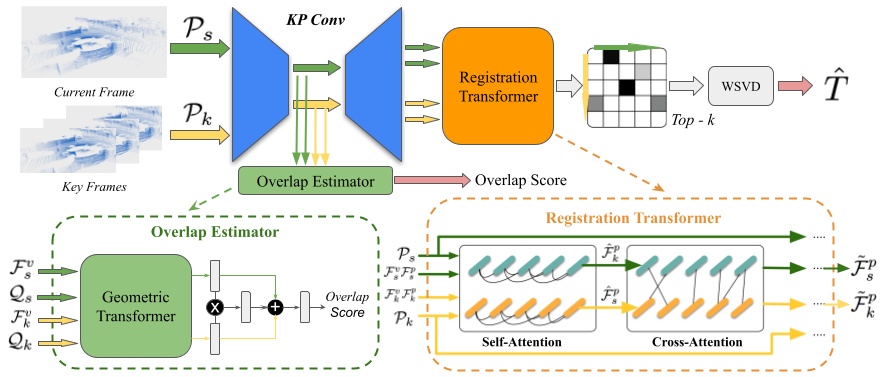}
    \caption{{\bf Overlap estimation and point registration pipeline}: For a given current frame and the keyframes recorded in the past, we use KPConv~\cite{thomas2019kpconv} to extract point and voxel features hierarchically. $\cp$ and $\cq$ are points and voxels, and $\cf$ represents features. $k$ and $s$ are historic frames and the current frame. We take the voxel features to estimate the overlap score by the Overlap Estimator. The voxel features are also used along with point features in estimating the relative poses in  the Registration Transformer. The Registration model uses attention heads to estimate the matching relationship between the points before using this matching relationship to compute the final pose.}
    \label{fig:pipeline}
\vspace{-1em}
\end{figure*}

\section{Approach}
In this section, we give the definition of the overlap and describe the details of the overlap and registration models, and finally we also provide the complete loop closure detection strategy.

In this work, the low-level odometry is achieved by Lio-sam~\cite{liosam}, which is based on GTSAM~\cite{dellaert2012factor}, where the map modification is done by the factor graph~\cite{dellaert2012factor} method. we assume the map rectification is well handled by Lio-sam and we only focus on the loop detection and registration. 

The observation input of our model is 3D Lidar point cloud, points $\p \in \RR^3$. The loop closure task includes: First, estimate the overlaps between the current keyframe $s$ and the selected previous keyframes, then select the potential matched historic frames; Second, predict the transformation matrix $T\in SE(3)$ between the current keyframe $s$ and the selected historic keyframe $k$. Then extract the best matches as closed loops and provide the transformation matrix as constraints for local odometry.

In this section, we give an overlap definition in \ref{sec:definition}, describe Overlap Estimator in \ref{sec:overlap} and Registration in \ref{sec:registration}. Finally introduce the whole pipeline of the loop detection in \ref{sec:loop_detection}


\subsection{{Overlap Definition}}
\label{sec:definition}

For overlap detection, a sequence of keyframes $k$ with points set $\cp_k$ is compared with the current frame $s$ of observation with points set $\cp_s$. For each frame, we  use a voxelization method~\cite{zhou2020end} to down-sample the point clouds corresponding to this frame. The position of each voxel is calculated by the mean value of all the points in the voxel using Equation \ref{eq:voxelization}:

\begin{align}
    \v_{s,i} = \frac{1}{|\cc_i|}\sum_{\p_m \in \cc_i} \p_m,
    \label{eq:voxelization}
\end{align}
where position $\v_{s,i}\in \cq_s$ is the position of voxel $i$ in the current keyframe  $s$, then we can also have $\v_{k,j} \in \cq_k$ as the position of voxel $j$ in one of the previous keyframes  $k$. $\cq_s$ and $\cq_k$ represent the voxel clouds {generated from point clouds $\cp_s$ and $\cp_k$} frame  $s$ and $k$, respectively. All the points in the voxel space of voxel $i$ compose the point set $\cc_i$, so $\p_m\in\RR^3$ represents the $m_\text{th}$ point in $\cc_i$. The overlap between the two voxel clouds of the current keyframe and a historic keyframe is given as

\begin{align}
    O(\cq_s,\cq_k) &= \frac{\sum_{i,j} o_{i, j} \ci_{i, j}}{\min\{\abs{\cq_s},\abs{\cq_k}\}}, \\
    \ci_{i, j} &= \left\{\begin{matrix}
1, &\text{ when }o_{i, j} > th\\ 
0, &\text{ otherwise}
\end{matrix}\right.
    \label{eq:overlap}
\end{align}
$\ci_{i, j}\in \{0, 1\}$ represents the indicator  function that gives 1 when two voxels $i$ and $j$ overlap, where $th$ is the threshold for the overlap score. In this project, we set $th=0.1$. Notice that $O(\cq_s,\cq_k)$ is not in $[0,1]$ because the denominator is the minimum number of two groups of voxels and one voxel can match several voxels from the other cloud. The $o_{i, j}$ represents the overlap score of one pair of voxels, $i, j$, and this is estimated by the Overlap Estimator in in Figure \ref{fig:pipeline}. Each voxel is a small point cloud, so the ground truth overlap score of each pair of voxels is calculated in a similar way to how we match two point clouds: we calculate the average Euclidean distance $\norm{\cdot}$ between two points clouds and then take the negative exponential value as the overlap score of the two voxels to ensure the values are [0-1] for easier training, as the follows: 
\begin{align}
    o_{i, j} = \exp{\frac{-\sum_{\p_m\in\cc_j}\norm{\p_m - \p_n}}{|\cc_j|}},
    \label{eq:vox_overlap}
\end{align}
where $\p_m\in\cc_j$ and $\p_n\in {\cc_T}_i$, $\cc_i$ and $\cc_j$ are point sets of voxels $i$ and $j$ in frames $s$ and $k$, and $|\cc_j|$ is the points number in set $\cc_j$. ${\cc_T}_i$ is the transformed point set of ${\cc_T}_i$ by the ground-truth transformation matrix $T$ (Figure~\ref{fig:pipeline}), and $\p_n$ is the nearest point in ${\cc_T}_i$ to $\p_m$. In our implementation, when the overlap score $O(\cq_s,\cq_k)> 0.5$, we consider the two keyframes overlapped.

\subsection{Overlap Estimation}
\label{sec:overlap}

Overlap detection detects overlapped point clouds. Because this step compares the current point cloud with multiple clouds in database, the computational cost is critical. Geometric Transformer's~\cite{geometric} pipeline is very computationally costly, but its Superpoint Matching Module encodes voxel's geometric information itself and also geometric relationship between two voxels in the voxel's features. Taking advantage of the geometric information, we design an Overlap Estimator to estimate the matched pairs, as Figure \ref{fig:pipeline}. To reduce the computation, we only use the voxels' features extracted by the encoder of the KPConv for the overlap estimation. The voxel features are put into the Superpoint Matching Module as Figure \ref{fig:pipeline} and then we can get processed voxel features (normalized), $\f^s$ and $\f^k$ for the current keyframe $s$ and one historic keyframe $k$. The feature similarity $\hat{d}()$ and overlap score $\hat{o}$ are:
\begin{align}
\hat{d}(\f_i^s, \f_j^k) &= 1 - \norm{\f_i^s - \f_j^k}  \in [0,1]\\
\hat{o}(\f_i^s, \f_j^k) &= a\left( g(h_s(\f_{i}^s) + h_c(\f_{i}^s \cdot \f_{j}^k) + h_k(\f_{j}^k)) \right)
\label{eq:feature_distance}
\end{align}
where $a(\cdot)$ is the Sigmoid activation function. $g(\cdot)$, $h_s(\cdot)$, $h_c(\cdot)$, and $h_k(\cdot)$ all contain a sequence of linear layers with ReLU activation function.  The details of the networks and feature dimensions are in Supplement Section A. ~\cite{appendix}. In the following equations, we use $o_{i,j}$ for the ground truth of the overlap score between the two voxels $i$ and $j$, $\hat{o}_{i,j}$ for $\hat{o}(\f_i^s, \f_j^k)$, and $\hat{d}_{i,j} $ for $\hat{d}(\f_i^s, \f_j^k)$.

The loss function of the whole overlap estimation model has two parts: the first part is the circle loss~\cite{sun2020circle} to learn a better matching between two voxel clouds and the second part is used for accurate overlap estimation.

As Equation \ref{eq:circle_loss}, $\cl_i$ is the total circle loss~\cite{sun2020circle} and the $\cl_l(s)$ and $\cl_l(k)$ represent the circle loss for the voxels in the current keyframe $s$ and the historic keyframe $k$, respectively. $\ci_{i,j}$ is defined in Equation \ref{eq:overlap}. Each $\cl_l$ contains loss functions $\cl_l^p$ and $\cl_l^n$, which are for overlapped pairs and non-overlapping pairs. $\cp$ and $\cn$ represent the set of overlapped and non-verlapped voxel pairs, respectiely. LSE is the LogSumExp loss to keep the loss numerically stable. $\alpha$ and $\beta$ are all hyper-parameters. 
\begin{align}
    \label{eq:circle_loss}
    \cl_i &= \cl_l(s) + \cl_l(k) \;\text{ where }\; \cl_l = \alpha \cl_l^p + \beta \cl_l^n \\
    \cl_l^p &= \frac{1}{\abs{\cp}} \text{LSE} (\ci_{i,j}(\hat{d}_{i,j} - \ci_{i,j})^2); \;(i,j)\in \cp  \\
    \cl_l^n &= \frac{1}{\abs{\cn}} \text{LSE} ((0 - \hat{d}_{i,j})^2); \;(i,j)\in \cn
\end{align}

The loss for the overlap values is $\cl_o$, where MSE represents the mean squared error and $\hat{O}=\{\hat{o}_{i,j}\}$ and $O=\{o_{i,j}\}$ are estimated and real overlap values for all voxels pairs, $\set{i, j}$, in the two frames $s$ and $k$:
\begin{align}
    \cl_o &= MSE(\hat{O} - O)
\end{align}
The total loss of the overlap estimation is $\cl_c = \cl_o + \cl_i$. Next,  we calculate the overlap of the matched frames based on  Equation \ref{eq:overlap}. If the keyframe and the current frame satisfy the overlapped criteria, those matched voxels are used for point registration.

\subsection{Point Registration}
\label{sec:registration}
\begin{figure}
    \centering
    \includegraphics[width=0.9\linewidth]{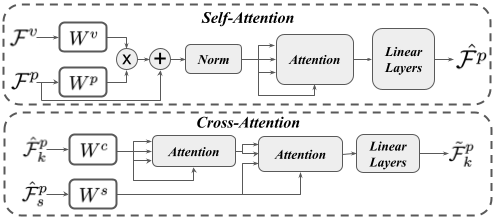}
    \caption{ {\bf Attention-based registration}: The registration model includes self-attention (up) and cross-attention (bottom) models in sequence.
    The voxel features $\cf^v$ and point features $\cf^p$ from two frames are inputs of the self-attention model and the cross-attention model takes the output of the self-attention model to process the relationship between two frames of points. This figure is an example of $\Tilde{\cf_k^p}$. }
    \label{fig:transformer}
\end{figure}

Registration is to provide an estimated transformation matrix for the potentially matched frames from the overlap estimation. Then the matrices are used to further select matched frames and are finally used as a constraint to low-level map rectification. In this section, we describe our novel method for point registration to generate the transformation matrix between two keyframes. Since we only use Lidar as the perceptive sensor, if the point clouds do not overlap much or are in very different shapes, it would be hard to estimate the accurate transformation matrix between them. Therefore, rather than using whole point clouds, we only choose the points from highly overlapped voxel pairs to leverage their matching relationships.

After the voxel matching step, we have the matched pairs of voxels $\{i, j\}$ with corresponding point sets $\cc_i$ and $\cc_j$ as the input of the registration step. As shown in Figure \ref{fig:pipeline}, the registration step includes a Registration Transformer model and a WSVD (weighted SVD) model~\cite{chen2017weighted}. The Registration transformer takes the features of voxels and points to get the score map of the points in two point clouds, and then the WSVD model takes the point pairs with the highest scores to generate the transformation matrix $\hat{T}$.

For the Registration Transformer, instead of only using the points in the registration task, as Figure \ref{fig:transformer} shows, we take the features of the matched voxels $\cf^v$ to provide the larger scale of (voxel) space information for the point matching, where $\cf^v=\set{\cf^v_s, \cf^v_k}$ and $\f_i^s\in \cf^v_s$ and $\f_j^k\in\cf^v_k$. As shown in Figure \ref{fig:transformer}, $\cf^p=\set{\cf^p_s,\cf^p_k}$ represents the features of points in both frames. We first project the processed voxel features, $W^v \cf^v$, to the space of processed point features, $W^p \cf^p$, where $W^v$ and $W^p$ are the weights for the features of voxels and points. After normalization, we put the features to the attention networks and then post-process the features by several linear layers. The entire process of the self-attention model  $k_{self}(\cdot)$ is on the left. Our attention model is based on the multi-head attention model~\cite{vaswani2017attention}. The output of the self-attention model is $\hat{\cf^p}$, as :
\begin{align}
    \hat{\cf^p} = k_{self} (\cf^v, \cf^p)
\end{align}

After we process both $s$ and $k$ frames by self-attention module and generate $\hat{\cf}_s^p$ and $\hat{\cf}_k^p$, the cross-attention model $k_{cross}(\cdot)$ takes them as inputs. $W^c$ and $W^s$ correspond to the weights of the inputs and the corresponding cross-attention output features of the points are $\Tilde{\cf^p_s}$ and $\Tilde{\cf^p_k}$. All these features have the shape of $v\times p\times d$, where $v$ is the size of matched voxels, $p$ is the size of points in each voxel, and $d$ is the feature dimension. 
\begin{align}
    \Tilde{\cf}_k^p = k_{cross}(\hat{\cf}_s^p, \hat{\cf}_k^p), \;\;\; \Tilde{\cf}_s^p = k_{cross}(\hat{\cf}_k^p, \hat{\cf}_s^p)
\end{align}

We use the Einstein sum to calculate the matching scores between two points features:

\begin{align}
    s[a, m, n] = \sum_{k\in \{0,...,d\}} \Tilde{\cf^p_k}[a,m,k] \times \Tilde{\cf^p_s}[a,n,k]
    \label{eq:points_score}
\end{align}
where $a\in[0, v)$ is the index of matched voxel pairs. $m\in[0, \abs{\cc_i})$ and $n\in[0, \abs{\cc_j})$ are the indices of the points from the point sets $\cc_i$ and $\cc_j$ of the overlapped voxels $i$ and $j$. Then we have the score map with the shape $(v, \abs{\cc_i}, \abs{\cc_j}))$. We choose the points with the biggest value ("Top-k" in Figure \ref{fig:pipeline} and $k=1$) in the last dimension to compose the point pairs with the score map $s$, which now has the shape of $(v, \abs{\cc_i})$. Those scores are used as weights for the point pairs to be put into the weighted SVD model, whose gradience functions can be found in the work~\cite{chen2017weighted}.

The loss function of the registration stage contains two parts: The first part aims at reducing the error of the estimated transformation matrix. Inspired by the transformation loss~\cite{wang2019deep}, the first part of the loss function is given as:
\[\cl_\text{matrix} = \norm{\hat{R}^TR_g - I}^2 + \norm{\hat{t}-t_g}^2\]
where $\hat{R}$ and $\hat{t}$ are the estimated rotation and translation transformations, and the estimated transformation matrix $\hat{T}$ is composed of $\hat{R}$ and $\hat{t}$. $I$ is a $(3\times 3)$ identity matrix. $R_g$ and $t_g$ are the ground truth transformations corresponding to rotation and translation.

Because the ground truth transformation matrices in some datasets~\cite{kitti, kitti360} are not very accurate, shown in Figure \ref{fig:inaccuracy} \cite{appendix}, we also need to provide some loss to compensate for the inaccuracy. The second part is aimed at reducing the distances between the two transformed point clouds $\hat{T} \p_s$ and $T_g \p_s$, where $\hat{T}$ and $T_g$ represent the estimated transformation matrix and the ground truth transformation matrix and $\p_s$ represents the points in the current keyframe:

\begin{align}
    \cl_\text{distance} = \text{MSE}(\hat{T} \p_s - T_g \p_s),
    \label{eq:mse_loss}
\end{align}

The total loss for the registration stage is given as:
\begin{align}
    \cl_r = \gamma \cl_\text{matrix} + \eta \cl_\text{distance},
\end{align}
where $\gamma$ and $\eta$ are coefficients. This loss function is for the registration of points. In the training phase, we combine the overlap loss function and registration loss function together $\cl=\cl_c +\cl_r$ for the whole training. 

\subsection{Loop Detection}
\label{sec:loop_detection}
\begin{algorithm}
\caption{{\bf Loop Detection}. The low level odometry is implemented by Lio-sam, and the following is the loop closure detection. $N=2$ in this project.
}\label{alg:rectification}
\begin{algorithmic}[1]
\Require {$\fr_c$ the current frame}
\Require {$\ck$ the set of previous keyframes in sequence}
\Procedure{Loop Closure}{}
\State ${\ck}_n \gets$ RemoveFarFrames($\ck, \fr_c$)
\For{$\fr_k \gets \ck_n$} 
\State $O(\co_s,\co_k) \gets$ OverlapEstimator
\If{$O(\co_s,\co_k) > O_{th}$}
\State $\hat{T} \gets$ PointRegistration($\cf_s^v, \cf_s^p,\cf_k^v, \cf_k^p$)
\If{$\hat{T}.translation < \d_{th}$}
\State PublishLoopConstraints
\EndIf
\EndIf
\EndFor
\State add $\fr_c$ to $\ck$ each N frames.
\EndProcedure
\end{algorithmic}
\end{algorithm}

As shown in Algorithm \ref{alg:rectification}, when a robot is moving, our model gets the frames from Lio-sam and saves each keyframe $\fr_c$ from every N received frames, where N is 2 in this project. 

The $RemoveFarFrames$ function removes the historic keyframes far from the current keyframe with a distance threshold $d_f$, because if the frames are too far it has less possibility as a loop. This function also removes the frames, $\ch_c$, in the last $d_h$ meters of the trajectory from the current keyframe $\fr_c$. For all saved keyframes $\fr_k$, we have the constraints:
\begin{align}
    \fr_k \notin \ch_c \;\;\text{and}\;\; \norm{\fr_k - \fr_c} < d_f
    \label{eq:distance_threshold}
\end{align}

The $OverlapEstimator$ predicts the overlap value between the two point clouds, and the $PointRegistration$ is the registration model, which estimates the transformation matrix between the two point clouds. These are two stages of filtering matched frames. The parameter values for different datasets can be different, and our values are in the Experiment Section. Given the transformation matrix, if the translation value is smaller than the threshold $\d_{th}=3$m, we consider the loop closed. The loop constraints are as what Lio-sam~\cite{liosam} takes and are published by $PublishLoopConstraints$ function.


\section{Results and Comparisons}
\label{sec:experiments}
We tested our method on KITTI~\cite{kitti} and KITTI-360~\cite{kitti360} datasets to validate our approach. We compared the accuracy of overlap estimation and pose estimation results with other state-of-the-art methods, including OverlapNet~\cite{overlapnet} and LCDNet~\cite{lcdnet}.  For more comparisons with different datasets (Nuscenes~\cite{nuscenes}, Complex Urban~\cite{urban}, NCLT~\cite{nclt} and MulRan~\cite{mulran}) are in Tables V-X in Supplement~\cite{appendix}.

The entire model is trained by the loop-closed data of all the datasets together (Nuscenes~\cite{nuscenes}, Complex Urban~\cite{urban}, NCLT~\cite{nclt} and MulRan~\cite{mulran}), without the pre-trained models of KPConv~\cite{thomas2019kpconv} or Geometric Transformer~\cite{geometric}. Specifically for the experiments, for KITTI and KITTI-260, we use Sequences [1,2,3,4,5,6,7,9,10] of KITTI and [0,3,4,5,6,7,10] of KITTI-360 for training and Sequences [0,8] of KITTI and [2,9] of KITTI-360 for evaluation. The Overlap Estimator and the Registration Transformer models are trained at the same time. 

\subsection{Runtime Analysis}

We compare the running time of our approach with other state-of-the-art approaches. For the experiments, we use an NVIDIA RTX A5000 GPU and an Intel Xeon(R) W-2255 CPU. The average running time of our method includes 5 parts: 1. Point processing of each pair of frames takes 0.5s with CPU or 0.2s with CUDA (around 0.1s for single frame); 2. Voxel feature processing costs 0.03s; 3. Point feature processing takes 0.001s; 4 Registration takes 0.016s. We can observe a computational bottleneck in the point preprocessing step, as we have to get neighbors of the points and voxels in 5-stage voxelization. We accelerated it using CUDA and the dynamic voxelization method~\cite{zhou2020end}.

Table \ref{tab:run_time_compare} compares different state-of-the-art methods. Our approach has a running time comparable  to these other approaches with similar accuracy. 


\begin{table}
    \centering
    \begin{tabular}{c | c | c | c | c } 
     \hline
     \multicolumn{5}{c}{Running Time of Learning-based Approaches (ms)} \\
     \hline
     RPMNet & DCP & LCDNet & Ours & Ours(CUDA) \\ 
     \hline
     854.79 & 456.18 &  1324.2 & 597 & 297 \\
     \hline
    \end{tabular}
    \caption{Our method is faster than most of the learning-based approaches in each processing.}
    \label{tab:run_time_compare}
\end{table}

\subsection{Loop Detection}
\begin{figure}
    \centering
    \includegraphics[width=\linewidth]{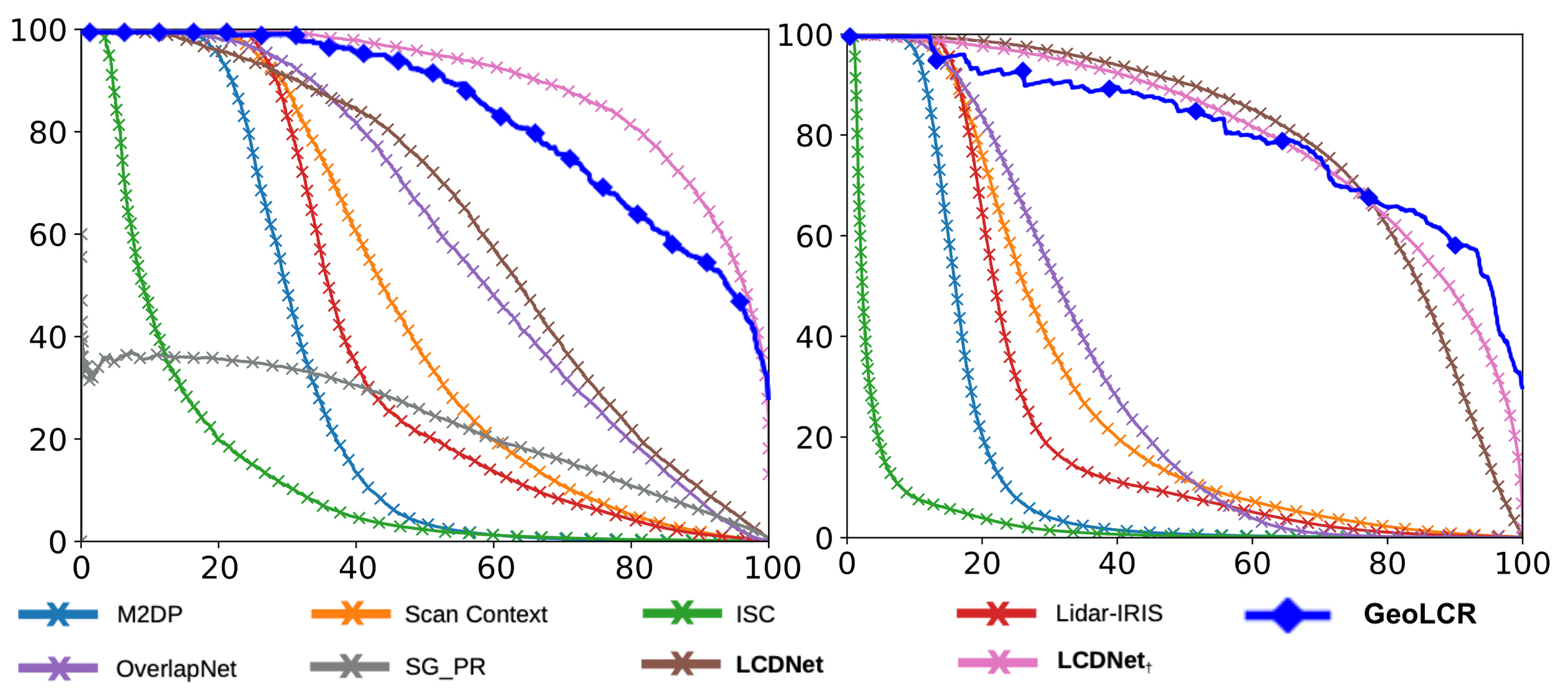}
    \caption{The recall precisions of the coarse matching results. the blue cubes represents our work (GeoLCR) and the pink line is LCDNet. We can see our coarse matching strategy has very competitive results with the LCDNet.}
    \label{fig:recall}
\end{figure}

Loop detection has two stages, the first stage is the overlap estimation, where we use Overlap Estimator to estimate the overlaps of point clouds. Because the overlap values are very related to frame distances as mentioned in Figure~\ref{fig:overlap_results}~\cite{appendix}, our false positive cases do not have very far distances. This makes it possible to use registration to filter negative matches. Therefore, the second stage is to use the registration to remove the far frames. 

We tested our Overlap Estimator model with the same critiera in LCDNet~\cite{lcdnet}. As Figure \ref{fig:recall}, our approach is the blue curve that has very competitive performance with LCDNet(pink curve). But instead of only having single stage place recognition(as LCDNet), after coarse matching we also use registration model to remove the far frames. Therefore, after estimating the transformation matrices, we remove the frames 3 meters away from the current frame as $\d_{th}=3m$ in \ref{sec:loop_detection}. As Table \ref{tab:coarse_fpfn}, after the registration step, there are mostly all matched frames cases left.

\begin{table}[]
    \centering
    \begin{tabular}{c|c|c}
    \hline
       Data        & False Postive & False Positive After Registration \\
       \hline
       KITTI 00    & 0.26          & 0.00 \\
       \hline
       KITI-360 09 & 0.22          & 0.01 \\
       \hline
    \end{tabular}
    \caption{The false positive and false negative of the loop closure detection models. After the registration we can get very accurate loops.}
    \label{tab:coarse_fpfn}
\end{table}

\subsection{Pose Estimation}
\begin{table*}[t]
\small
\centering 

\begin{threeparttable}

\begin{tabular}{c | ccc | ccc}
 & \multicolumn{3}{c|}{Seq. 00} & \multicolumn{3}{c}{Seq. 08} \\

\hline
& Success & TE (m) & RE (deg) & Success & TE (m)  & RE (deg) \\
&  & (succ. / all) & (succ. / all) &  &  (succ. / all) & (succ. / all) \\ 

\hline
GICP~\cite{gicp} & - & - / 1.38 & - / 6.22 & - & - / 3.71 & - / 153.7 \\
\hline
TEASER~\cite{teaser} & 34.06\% &0.98 / 2.72 & 1.33 / 15.85 & 17.13\% &1.34 / 3.83& 1.93 / 29.19\\
\hline
\hline
OverlapNet~\cite{overlapnet} & 83.86\% & - / - & 1.28 / 3.89 & 0.10\% & - / - & 2.03 / 65.45 \\


\hline
DCP~\cite{wang2019deep} & 50.71\% & 0.98 / 1.83 & 1.14 / 6.61 & 0\% & - / 4.01 & - / 161.24 \\

 
\hline
LCDNet~\cite{lcdnet} & \textbf{100\%} & {0.11 / 0.11} & \textbf{0.12 / 0.12} & \textbf{100\%} & {0.15 / 0.15} & \textbf{0.34 / 0.34} \\

\hline
Geometric Transformer~\cite{geometric} & - & - / 0.08  & - / 0.32  & - & - / 0.13  & - / 0.67  \\
\hline
\hline
Ours (GeoLCR: wo Coarse ft) & \textbf{100\%} & \textbf{0.07 / 0.07} & \underline{0.43 / 0.43} & \textbf{100\%} & \textbf{0.12 / 0.12} & \underline{0.51 / 0.51} \\

\hline
Ours (GeoLCR: w Coarse ft) & \textbf{100\%} & \textbf{0.06 / 0.06} & 0.33 / 0.33 & \textbf{100\%} & \textbf{0.11 / 0.11} & {0.47 / 0.47} \\
\hline

\end{tabular}
\end{threeparttable}
\caption{{\bf Pose estimation error for positive pairs in the KITTI dataset}. TE and RE are Translation and Rotation Errors. In both 0 and 8 sequences, our approach achieves better performance in translation accuracy and comparable rotation accuracy. In both cases, our method succeeded in detecting all the loop closure frames. (succ./all) means the errors for the frame pairs in successfully detected loops and in all loops in the database.}
\label{table:pose-estimation-comparison-kitti}
\end{table*}

\begin{table*}[t]
\small
\centering 

\begin{threeparttable}

\begin{tabular}{c | ccc | ccc}
 & \multicolumn{3}{c|}{Seq. 02} & \multicolumn{3}{c}{Seq. 09} \\
 
\hline
& Success & TE (m) & RE (deg) & Success & TE (m)  & RE (deg) \\
&  & (succ. / all) & (succ. / all) &  &  (succ. / all) & (succ. / all) \\ 

\hline
GICP~\cite{gicp} & - & - / 3.51 & - / 138.61 & - & - / 2.39 & - / 79.72 \\
\hline
TEASER~\cite{teaser} &27.02\% & 1.25 / 3.16 & 1.83 / 19.16 & 30.32\% & 1.14 / 2.91 & 1.46 / 19.22 \\
\hline
\hline

OverlapNet & 11.42\% & - / - & 1.79 / 76.74 & 54.33\% & - / - & 1.38 / 33.62 \\


\hline

DCP & 5.62\% & 1.09 / 3.14 & 1.36 / 149.27 & 30.10\% & 1.04 / 2.30 & 1.06 / 64.86 \\


 


\hline

LCDNet & 98.55\% & 0.27 / 0.32 & 0.32 / 0.34 & \textbf{100\%} & 0.20 / 0.20 & 0.22 / 0.22 \\
\hline
Geometric Transformer & - & - / 0.27 & - / 1.46 & - & - / 0.12 & - / 0.41 \\
\hline
\hline

Ours (GeoLCR: wo Voxel ft) & \textbf{100\%} & \textbf{0.14 / 0.14} & \underline{0.42 / 0.42} & \textbf{100\%} & \textbf{0.10 / 0.10} & \underline{0.32 / 0.32} \\

\hline

Ours (GeoLCR: w Voxel ft) & \textbf{100\%} & \textbf{0.08 / 0.08} & \underline{0.35 / 0.35} & \textbf{100\%} & \textbf{0.08 / 0.08} & \underline{0.26 / 0.26} \\

\hline

\end{tabular}
\end{threeparttable}
\caption{{\bf Pose estimation error for positive pairs in the KITTI-360 dataset}. For both of the two sequences in the KITTI-360, our method achieved far better results in estimating translation(TE: Translation Error). On both Sequence 02 and 09, our method estimates poses up to {\bf two times better} than other algorithms in translation and shows comparable results in rotation(RE: Rotation Error). Also, our method succeeded in detecting loop closures for both the sequences (i.e. 100\% success rate), while every other method failed on Sequence 02.}
\label{table:pose-estimation-comparison-kitti360}
\vspace{-2em}
\end{table*}

To demonstrate the accuracy of the prediction of transformation matrices, we evaluated our model by the metrics of LCD-Net~\cite{lcdnet} in KITTI and KITTI-360 datasets. We compared our registration transformer's performance with other algorithms. Since this module is used for loop-closing frames, we have to temporarily select  the distant frames that closely overlap. To make a fair comparison with the other algorithms, we used the same criteria as LCDNet~\cite{lcdnet} to choose overlapped frames to test registration performance.

In Tables~\ref{table:pose-estimation-comparison-kitti} and ~\ref{table:pose-estimation-comparison-kitti360}, we present pose estimation errors for two sequences in each of the datasets. We compared both classical methods and learning-based methods, where LCDNet~\cite{lcdnet} is the closest competitor. Lio-Sam~\cite{liosam} uses ICP, but GICP is better~\cite{gicp} in registration so we compare with GICP. The translation error (TE) is calculated by the Euclidean distance between the two frames. The rotation error (RE) is calculated as: 
\[RE = \arccos{\frac{trace(\hat{R}^T R_\text{g}) - 1.0}{2}}\]
where $\hat{R}$ is the estimated rotation matrix and $R_\text{g}$ is the ground truth rotation matrix. 
We also did ablation study that compares the models w/wo the coarse features. From the tables~\ref{table:pose-estimation-comparison-kitti} and ~\ref{table:pose-estimation-comparison-kitti360}, voxels features help in both translation and rotation estimations. Comparing our model without Voxels' features with Geometric Transformer~\cite{geometric}, we can see the Registration Transformer in our model does better than the Sinkhorn algorithm used in Geometric Transformer~\cite{geometric}. This also demonstrates the power of transformers in cross relation estimation.

In Tables~\ref{table:pose-estimation-comparison-kitti} and \ref{table:pose-estimation-comparison-kitti360}, we can observe that our Registration Transformer achieves the best results in translation estimation and comparable results in rotation estimation. Before doing SVD, compared with LCDNet, our approach uses transformers to enhance the point features by the other frame as Figure~\ref{fig:pipeline}, which contributes to the more accurate SVD results.  Compared to other methods, our approach has at least a $30\%$ improvement in translation estimation. For the rotation estimation, our approach is comparable to LCDNet. But because our model is trained by different datasets, we are more generalized to different sorts of point clouds as shown in Tables V-X in ~\cite{appendix}.

\section{Limitations, Conclusions, and Future Work}
We present a new learning pipeline, GeoLCR,  for loop closure detection and points registration. We compare the accuracy of GeoLCR with LCDNet, OverlapNet, and other approaches in both KITTI and KITTI-360 datasets for points registration. Our method achieves up to {\bf two times} improvements in terms of translation accuracy, and we also achieve comparable rotation accuracy. 

Our approach has some limitations. First, our current formulation is only designed for using the 3D Lidar as the perception sensor. In practice, Lidar can be sensitive to rainy or foggy weather. In future work, RGB data can help to improve the robustness of the system. Second, in this work, we use voxels to remove redundant points, but the method still uses many points for registration. In the future, we can choose points directly from the point level for registration.

\bibliographystyle{IEEEtran}
\bibliography{references}

\newpage
\section{Supplement}

\begin{figure}
    \centering
\includegraphics[width=\linewidth]{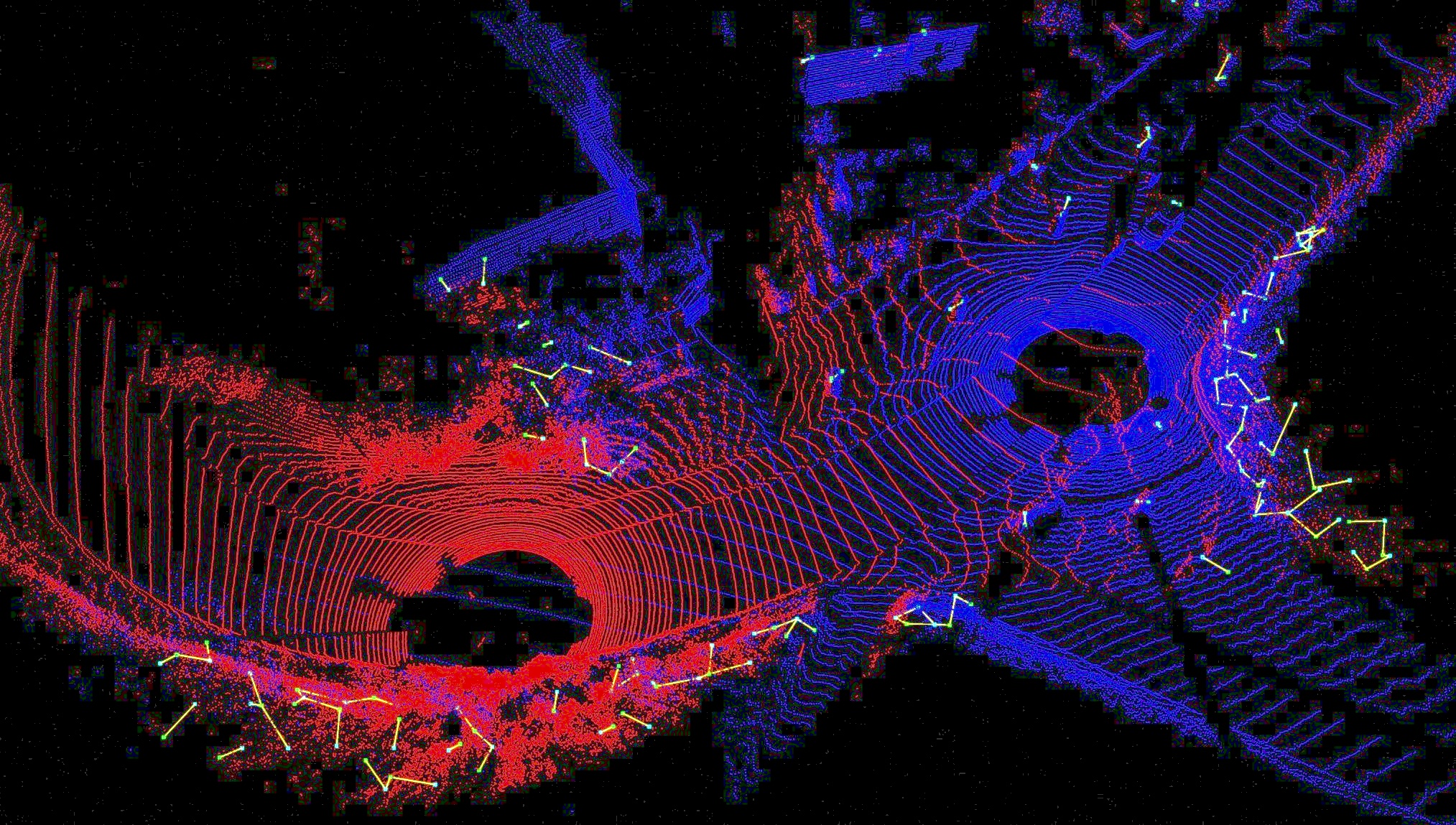}
    \caption{An example of an overlapping area, where the red and blue points are from two different point clouds. The small green cubes are the centers of the voxels. The matched voxels are connected by the yellow lines. As shown, the matched voxels are mostly from the two sides of the road, which have more features than the road.}
    \label{fig:overlap_figure}
\vspace{-1em}
\end{figure}

\subsection{Network Structures}

\begin{figure*}
    \centering
    \includegraphics[width=0.8\linewidth]{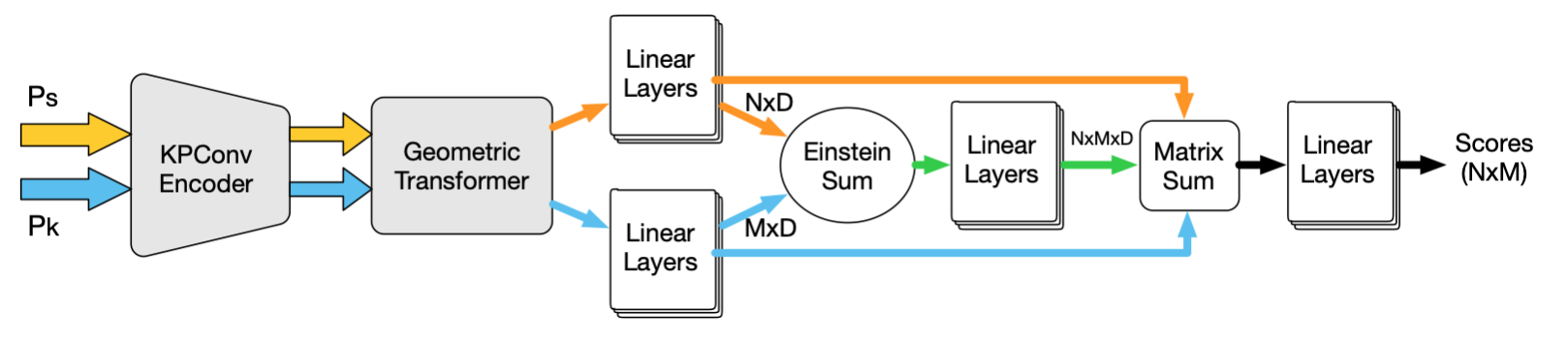}
    \caption{Overlap Estimator: This is the network structure of the Overlap Estimator, which is composed of a  KPConv encoder, Geometric Transformer, and Overlap Model.}
    \label{fig:overlap_networks}
\end{figure*}

In Figure \ref{fig:overlap_networks}, the inputs are points, $\cp_s$ and $\cp_k$, from two key frames, $s$ and $k$. After KPConv~\cite{thomas2019kpconv} feature encoder and Geometric Transformer~\cite{geometric}, we process the features by 2 consecutive linear layers, and the details of the layers are mentioned in Section \ref{sec:overlap}. The outputs of the features are $N\times C$ and $M\times C$, respectively. Then we use Einstein sum to project the features to $N\times M\times C $ space. Then after post feature process, we project the two previous  voxel features to the $N\times M\times C $ space again for more information about the voxel features themselves. The final output is the score map in the $N\times M$ space. The voxels features $\f^s$ and $\f^k$ have dimension of 256. For the linear layers, $h_s(\cdot)$, $h_c(\cdot)$, and $h_k(\cdot)$, we found two linear layers with the same feature dimensions(512) are good enough and $g(\cdot)$ outputs only one value for each pair as the estimation.

\begin{figure*}
    \centering
    \includegraphics[width=0.8\linewidth]{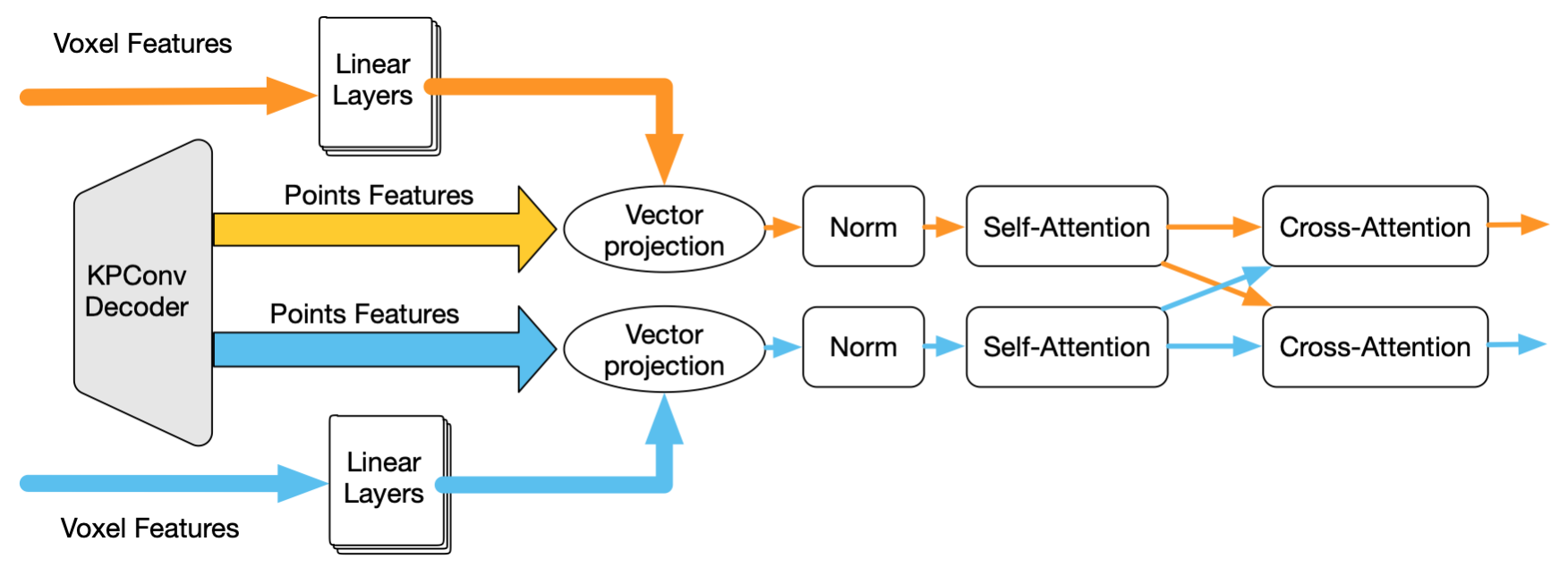}
    \caption{Register: This is the network structure of the transformer model of the registration process. It takes the voxel features and point features and process the features by the attention models.}
    \label{fig:attention_networks}
\end{figure*}

In Figure \ref{fig:attention_networks}, we leverage the coarse features and project the point features to the voxel feature space. After the normalization, we use two attention models to process the self relationship in each point cloud and cross relationship between two point clouds, then output the point features for point matching. See Section 3.3 for the details of the Self-Attention and Cross-Attention models.

For the loss function of the overlap estimation, the circle loss has the same values (0.5) for $\alpha$ and $\beta$. The registration stage has $\gamma$ and $\eta$ are all $1.0$.

\subsection{Coarse Matching Analysis}
The overlap values are calculated by the Equation \ref{eq:overlap}. After choosing the threshold $th$, we have the matching pairs $\set{i, j}$. Figure \ref{fig:overlap_figure} shows the voxels matching after registration, where the green dots are voxels and the yellow lines indicates the matches voxel pairs from the two point clouds. We also show the matching in the video of the project.

\begin{figure}
    \centering
    \includegraphics[width=\linewidth]{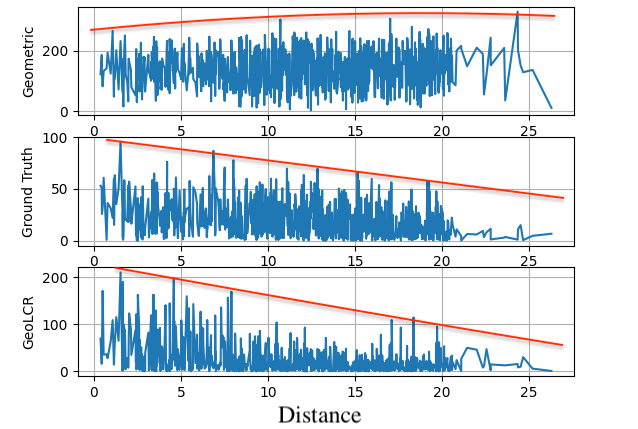}
    \caption{Overlap Estimation: The sub-figures are overlapped results tested in KITTI Sequence 7 with Geometric Transformer~\cite{geometric} (Top), Ground Truth transformations (Middle), and our approach  (Bottom). Y-axis is the overlap score and X-axis is the distance. The orange curves sketch the trend of the three curves (not computed by any specific values, only draw to indicate the trend). With the distance increasing, our approach is very effective for estimating the overlap values of the point clouds. It is also easy to choose a threshold value to the overlapping score for loop closure detection.}
    \label{fig:overlap_results}

\vspace{-1em}
\end{figure}

\begin{table*}[]
\small
\centering 
\begin{threeparttable}
\begin{tabular}{c | cc | cc |cc }
Mulran~\cite{mulran} & \multicolumn{2}{c|}{ Sejong01 } & \multicolumn{2}{c|}{ Sejong02 } & \multicolumn{2}{c}{ Sejong03 } \\
\hline
& TE (m)        & RE (deg)      & TE (m)         & RE (deg)     & TE (m)         & RE (deg)   \\
\hline
Ours (GeoLCR) & 0.47 & 0.78 & 0.44 & 0.73 & 0.74 & 1.12  \\
\hline
LCDNet~\cite{lcdnet} & 5.44 & 20.81 & 4.42 & 16.57 & 5.26 & 19.59  \\
\hline
Geometric Transformer~\cite{geometric} & 0.14 & 0.38 & 0.15 & 0.42 & 0.41 & 0.30  \\
\hline
GICP~\cite{gicp} & 0.89 & 0.86 & 1.02 & 0.33 & 1.64 & 1.07  \\
\hline
\end{tabular}
\end{threeparttable}
\caption{{\bf Pose estimation error in the Mulran dataset~\cite{mulran}}. }
\label{tb:sup_mulran}
\end{table*}

\begin{table*}[]
\small
\centering 
\begin{threeparttable}
\begin{tabular}{c | cc | cc  }
Complex Urban~\cite{urban} & \multicolumn{2}{c|}{ Urban07 } & \multicolumn{2}{c}{ Urban08 } \\
\hline
& TE (m)        & RE (deg)    & TE (m)         & RE (deg)      \\

\hline
Ours (GeoLCR) & 0.71 & 3.12 & 0.65 & 2.50  \\
\hline
LCDNet~\cite{lcdnet} & 4.18 & 54.92 & 3.64 & 37.84\\
\hline
Geometric Transformer~\cite{geometric} & 0.87 & 6.61 & 1.28 & 8.97 \\
\hline
GICP~\cite{gicp} & 2.13 & 83.23 & 1.44 & 29.00 \\
\hline
\end{tabular}
\end{threeparttable}
\caption{{\bf Pose estimation error in the Complex Urban dataset}. }
\label{tb:sup_urban}
\end{table*}

\begin{table*}[]
\small
\centering 
\begin{threeparttable}
\begin{tabular}{c | cc | cc | cc  }
NCLT~\cite{nclt} & \multicolumn{2}{c|}{ 2012-08-20 } & \multicolumn{2}{c|}{ 2012-10-28 } & \multicolumn{2}{c}{ 2012-11-16 } \\
\hline
& TE (m)        & RE (deg)    & TE (m)         & RE (deg)  & TE (m)         & RE (deg)      \\
\hline
Ours (GeoLCR) & 1.24 & 10.1 & 1.63 & 15.1 & 1.64  & 13.9 \\
\hline
LCDNet~\cite{lcdnet} & - & - & - & - & -  & - \\
\hline
Geometric Transformer~\cite{geometric} & 1.39 & 10.23 & 2.42 & 17.00 & 2.29  & 17.12 \\
\hline
GICP~\cite{gicp} & 4.17 & 128.98 & 4.05 & 107.83 & 3.37  & 86.69 \\
\hline
\end{tabular}
\end{threeparttable}
\caption{{\bf Pose estimation error in the NCLT dataset}. Because the dataset is very sparse and there are not many points(16-channel Lidar), the LCD-Net failed in the dataset. For other methods, because there are some places in open space, there are not many features around the robot, all the algorithms perform bad in the situation, thus the average performance in NCLT dataset is generally worse than other datasets.}
\label{tb:sup_nclt}
\end{table*}

We compare the overlap values directly from Superpoint Matching Module~\cite{geometric}, Ground truth, and our approach. The LCDNet~\cite{lcdnet} is not here because it is trained with some positive/negative overlapped cases, not distances. The overlap values of the two point clouds are calculated by Equation \ref{eq:overlap}. Also, as the Superpoint Matching Module is not able to directly estimate the overlap values of two point clouds, we took the exponential negative feature distances as the overlap values  $o_{i,j}$ of voxels pairs, and the ground-truth voxel overlap values $o_{i,j}$ are computed by Equation \ref{eq:vox_overlap}. For our approach, the voxel overlap values $o_{i,j}$ are directly computed from the overlap estimation we presented before, and the overlap of the two point clouds is also calculated by  \ref{eq:overlap}.

From Figure \ref{fig:overlap_results}, we can see that the overlap values of Superpoint Matching Module~\cite{geometric} cannot be used for overlap detection. In contrast, the overlap values from our model align well with the distance between two point clouds. Therefore, we can easily use it  for loop closure estimation by giving a threshold to the overlap values. 

\subsection{Comparisons with Other Datasets}
In this section, we compare our method with LCD-Net~\cite{lcdnet} and Geometric Transformer~\cite{geometric} in NCLT, Complex Urban, and MulRan datasets in terms of registration errors. The errors are collected averagely among all loop-closed conditions in the sequence. From the tables \ref{tb:sup_mulran} and \ref{tb:sup_urban}, we can see our approach has a comparable results w.r.t. Geometric Transformer~\cite{geometric} and better results than GICP~\cite{gicp} and LCD-Net~\cite{lcdnet}. For the Table \ref{tb:sup_nclt}, the NCLT~\cite{nclt} dataset is collected from small robots with 16-channel Lidar, which has very sparse point and less number of points than other datasets. Therefore, all methods have bad performance in this datasets because there are some cases in open space with less features of the objects and in this condition all methods would perform bad. However, compared with other methods our method still achieved good average accuracy of both rotation and translation.

\subsection{SLAM Trajectory}
\begin{figure}
    \centering
    \includegraphics[width=0.8\linewidth]{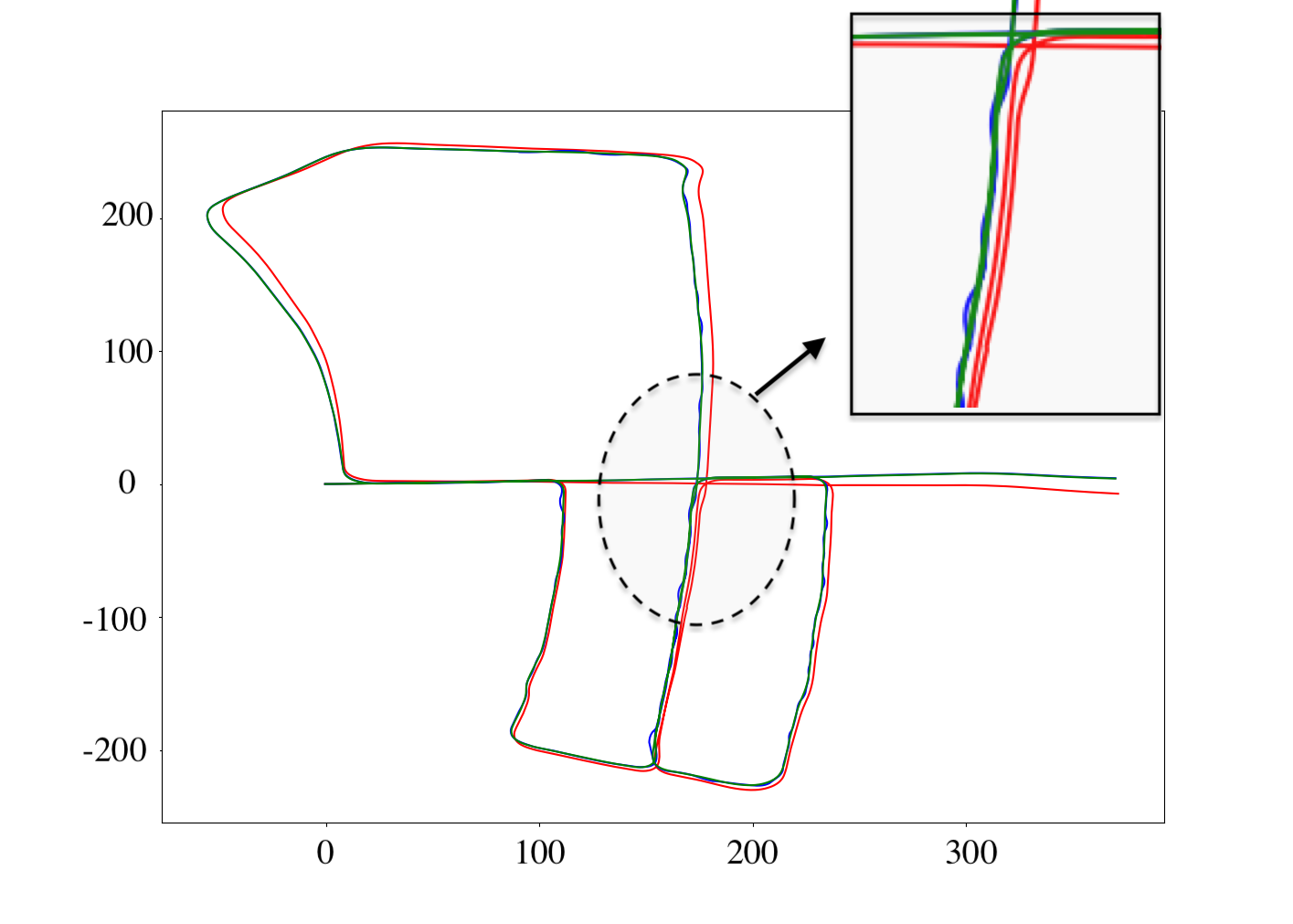}
    \caption{The x and y axes are all meters. This figure shows the result of GeoLCR (Blue) compared with GICP (Red) and the ground truth (Green) trajectories. Because our method provides more accurate transformation matrices and loop closed detection, our approach achieves $0.26m$ error of the trajectory less than the $83.6m$ by GICP in Sequence 05 of the KITTI dataset.
    }
    \label{fig:map_rectification}
\vspace{-1em}
\end{figure}

Lio-sam uses ICP to estimate the transformation matrix for loop closure constraints. In the experiment, we use a better GICP~\cite{gicp} to substitute the ICP and compare it with our approach. Because the Lio-sam algorithm is relatively accurate, the distance threshold $d_f$ in Equation \ref{eq:distance_threshold} is set as 5 meters and the $d_h$ is set to 20 meters in this experiment for the KITTI sequence dataset. For more comparisons please refer to the supplement material~\cite{appendix}. The keyframes of the loop closure models are chosen from the frames published by Lio-sam odometry as  Algorithm \ref{alg:rectification}. Figure \ref{fig:map_rectification} shows a sample of the complete trajectory using KITTI dataset.


\begin{figure}
    \centering
    \includegraphics[width=0.49\linewidth]{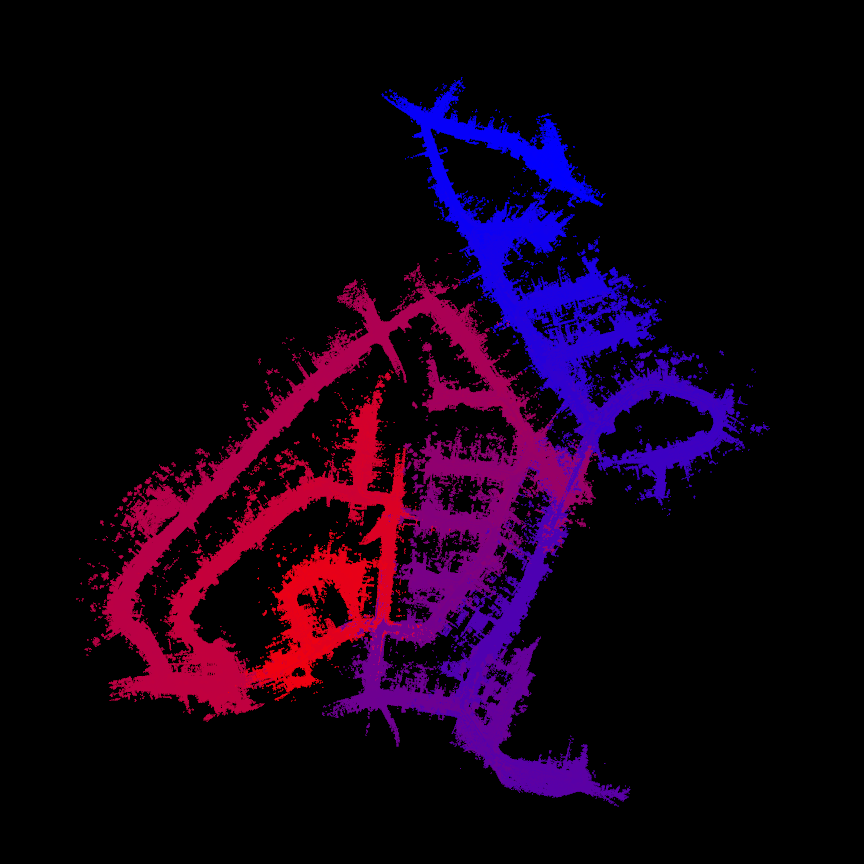}
    \includegraphics[width=0.49\linewidth]{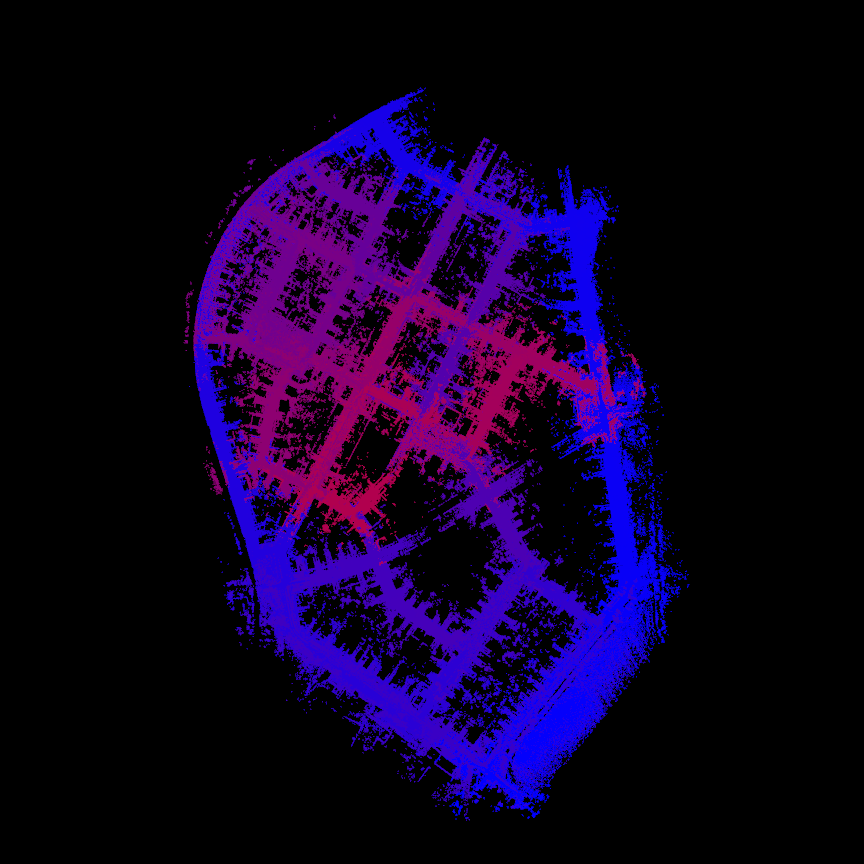}
    \caption{Trajectory of Sequence 02 (left) and 09 (right) in  the KITTI-360 dataset. The color coding indicates the starting point (Blue) and the end point (Red) of the moving vehicle. Note that loop closures occur at the points where two different colors overlap. Our method succeeded in detecting all of the loop closures in these sequences by using our novel point registration transformer, while other previous methods failed on these sequences.}
    \label{fig:teaser}
\vspace{-2em}
\end{figure}

Further implementations are shown in Figure \ref{fig:teaser}. The built maps are KITI-360 sequence 02 and 09, where blue points are starting points and red points are ending points.

\subsection{Comparisons with All Near Frames}
6
We compared with LCD-Net~\cite{lcdnet} in other KITTI, KITTI-360, and Nuscenes datasets with not only loop closed cases, but all frame pairs with near distance. The table shows the results from all sequences from KITTI-360, Sequences 00 and 08 from KITTI datasets, and the results from the Nuscenes dataset. Compared with LCD-Net, our approach has a higher rate of nearby-frame detection and fewer translation and rotation errors. From Table \ref{tb:sup_kitti}-\ref{tb:sup_nuscenes}, we observe that our approach can detect all the near frames cases. Furthermore, our registration is more accurate than LCD-Net in both translation and rotation estimations.

For the two KITTI sequences in Table \ref{tb:sup_kitti}, we directly use the results from the table of LCD-NET. Our approach achieves comparable results w.r.t. translation and rotation errors.

\begin{figure*}
    \centering
    \includegraphics[width=0.45\linewidth,height=0.3\linewidth]{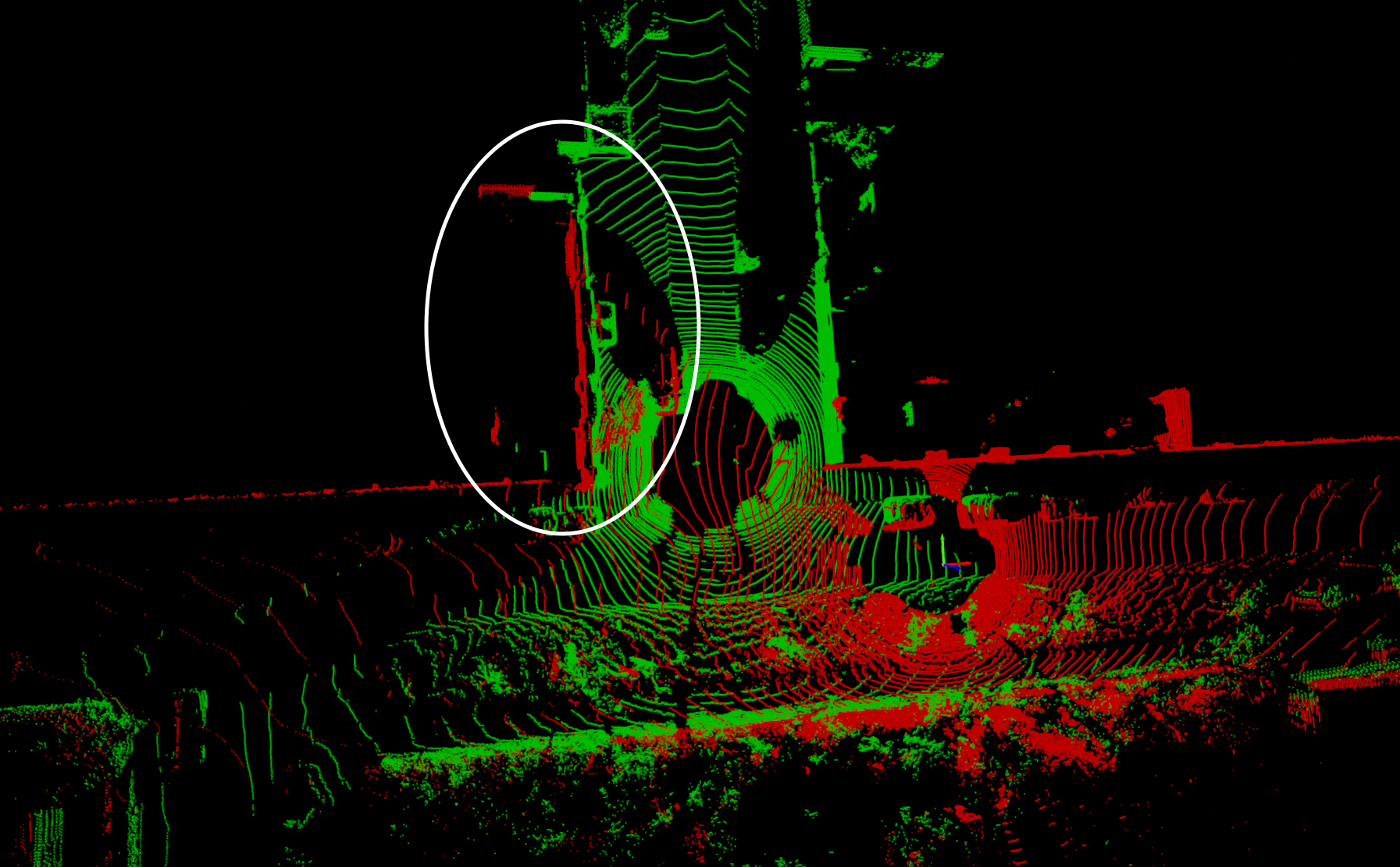}
    \includegraphics[width=0.45\linewidth,height=0.3\linewidth]{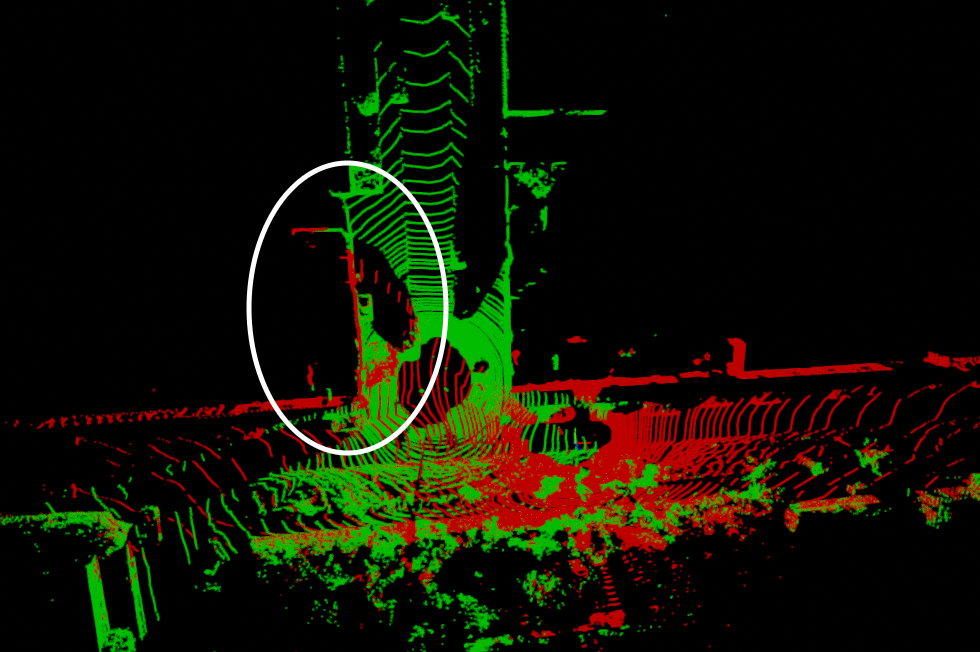}
    \caption{KITTI dataset Sequence 00. The white circle of the left figure shows the two point clouds are not well aligned by the ground truth transformation matrix. The white circle of the right figure shows our result of the estimated matrix.}
    \label{fig:inaccuracy}
\end{figure*}

\begin{table*}[]
\small
\centering 

\begin{threeparttable}

\begin{tabular}{c | ccc | ccc }
 & \multicolumn{3}{c|}{Ours (GeoLCR)} & \multicolumn{3}{c}{LCDNet~\cite{lcdnet}} \\
 
\hline
& Success & TE (m) & RE (deg) & Success & TE (m)  & RE (deg) \\
&  & (succ. / all) & (succ. / all) &  &  (succ. / all) & (succ. / all) \\

\hline
KITTI Seq 00 & 
\textbf{100\%} & \textbf{0.11 / 0.11} & \textbf{0.12 / 0.12} &
\textbf{100\%} & 0.14 / 0.14 & {0.13 / 0.13} \\

\hline
KITTI Seq 08& 
\textbf{100\%} & \textbf{0.15 / 0.15} & 0.34 / 0.34  & 
\textbf{100\%} & 0.19 / 0.19 & \textbf{0.24 / 0.24} \\

\hline
\end{tabular}
\begin{tablenotes}
    \item [1]: We denote the best result with bold face.
\end{tablenotes}
\end{threeparttable}
\caption{{\bf Pose estimation error for positive pairs in the  KITTI dataset}. Our method achieved a  comparable result for translation and better performance for rotation.}
\label{tb:sup_kitti}
\end{table*}

For the KITTI-360 sequences, LCD-Net is tested by using the open-sourced pre-trained model. To achieve the best performance of LCD-Net, we also applied ground point removal according to the Github~\cite{lcdnet} command lines. Table \ref{tb:sup_kitti360} shows that our method is very robust in nearby frames detection, which is always $100\%$ compared with the LCD-Net (around 50\%). Our approach is also much more accurate in nearby-frame detection and can achieve much higher accuracy than LCD-Net in translation and rotation. 

\begin{table*}[]
\small
\centering 

\begin{threeparttable}

\begin{tabular}{c | ccc | ccc }
 & \multicolumn{3}{c|}{Ours (GeoLCR)} & \multicolumn{3}{c}{LCDNet~\cite{lcdnet}} \\
 
\hline
& Success & TE (m) & RE (deg) & Success & TE (m)  & RE (deg) \\
&  & (succ. / all) & (succ. / all) &  &  (succ. / all) & (succ. / all) \\ 

\hline
KITTI-360 Seq 0& 
\textbf{100\%} & \textbf{0.14 / 0.14} & 0.13 / 0.13  & 
{44\%} & 0.97 /2.0  & 1.12 /27.75 \\

\hline
KITTI-360 Seq 2& 
\textbf{100\%} & \textbf{ 0.17 / 0.17} & 0.18 / 0.18  & 
{49.8\%} & 1.00 / 1.93  & 1.37 / 14.98 \\

\hline
KITTI-360 Seq 4& 
\textbf{100\%} & \textbf{0.16 / 0.16 } & 0.15 / 0.15  & 
{44\%} & 1.00 / 2.05  & 1.19 / 20.01 \\
 
\hline
KITTI-360 Seq 5& 
\textbf{100\%} & \textbf{0.16 / 0.16} & 0.15 / 0.15  & 
{47\%} & 0.97 / 2.03 & 1.39 / 22.1\\

\hline
KITTI-360 Seq 6& 
\textbf{100\%} & \textbf{0.17 / 0.17} & 0.15 / 0.15  & 
{57\%} & 0.95 /1.68  & 1.21 /14.70 \\
\hline
KITTI-360 Seq 9& 
\textbf{100\%} & \textbf{0.14 / 0.14} & 0.12 / 0.12  & 
{57\%} & 0.92 / 1.54  & 1.13 / 9.82 \\

\hline

\end{tabular}
\begin{tablenotes}
    \item [1]: We denote the best result with bold face.
\end{tablenotes}
\end{threeparttable}
\caption{{\bf Pose estimation error for positive pairs in the KITTI-360 dataset}. Our approach achieves higher accuracy in rotation and translation estimations and more robust accuracy in nearby-frame detection.}
\label{tb:sup_kitti360}
\end{table*}

The Nuscenes dataset has more features than KITTI. There are more pedestrians, moving cars, and buildings than in the scenarios in KITTI or KITTI-360. we collect the nearby-frame pair using a similar method and test the result with the dataset. From Table \ref{tb:sup_nuscenes}, we can see LCD-Net does not perform well in the nearby-frame situations in the Nuscenes dataset and the success rate is only $21\%$. However, our approach is still robust in the Nuscenes dataset with $100\%$ success rate and accurate rotation and translation estimations.

\begin{table*}[]
\small
\centering 

\begin{threeparttable}

\begin{tabular}{c | ccc  }

\hline
Nuscenes & Success & TE (m) (succ. / all) & RE (deg) (succ. / all) \\
 
\hline
Ours (GeoLCR)& 
\textbf{100\%} & \textbf{0.15 / 0.15} & 0.12 / 0.12  \\

\hline
LCDNet~\cite{lcdnet}& 
{21\%} & 0.57 /11.59  & 1.91 /20.95 \\
\hline
\end{tabular}
\begin{tablenotes}
    \item [1]: We denote the best result with bold face.
\end{tablenotes}
\end{threeparttable}
\caption{{\bf Pose estimation error for positive pairs in the Nuscenes dataset}. The Nuscenes dataset, with more dynamic obstacles, is more complex than the KITTI or the KITTI-360. Our method achieved a comparable result for translation and better performance for rotation.}
\label{tb:sup_nuscenes}
\end{table*}

\subsection{Inaccurate Ground Truth}
The public datasets do not always have accurate ground truth transformation matrices. As Figure \ref{fig:inaccuracy}, the red points are transformed by the ground truth transformation matrix. The white circle shows the two point clouds are not aligned only by using the ground truth transformation matrix. Therefore, in our loss function \ref{eq:mse_loss}, we also use the distance of the points to optimize the estimated transformation matrices.

\end{document}